\documentclass[journal]{IEEEtran}
\usepackage{amsmath,bm}
\usepackage{amsfonts,amssymb}
\usepackage{graphicx}
\usepackage{caption}
\usepackage{subfigure}

\usepackage{subfloat}

\usepackage{booktabs}
\usepackage{multirow}
\usepackage{color}
\usepackage{algorithm}
\usepackage{algorithmic}
\usepackage[switch]{{lineno}}
\usepackage{colortbl}
\definecolor{mygray}{gray}{.9}
\usepackage{dsfont}
\usepackage{tabu}

\label{definition}
\def\htheta{{h_{\bm{\theta}}}}
\def\bx{{\bm{x}}}
\def\bw{{\bm{w}}}
\def\btheta{{\bm{\theta}}}
\def\bp{{\bm{p}}}

\def\bz{{\bm{z}}}

\newtheorem{proposition}{Proposition}

\usepackage{xcolor}
\usepackage{xpatch}
\makeatletter
\def\changeBibColor#1{%
	\in@{#1}{}
	\ifin@\color{red}\else\normalcolor\fi	
}
\xpatchcmd\@bibitem
{\item}
{\changeBibColor{#1}\item}
{}{\fail}
\xpatchcmd\@lbibitem
{\item}
{\changeBibColor{#2}\item}
{}{\fail}
\makeatother

\usepackage{cite}

\ifCLASSINFOpdf
\else
\fi
\begin{document}
%
\title{InfoAT: Improving Adversarial Training Using the Information Bottleneck Principle}
%
%
%

\author{Mengting~Xu, Tao Zhang, Zhongnian Li,
        and Daoqiang Zhang
\thanks{ Mengting Xu, Tao Zhang, Zhongnian Li and Daoqiang Zhang are with the College of Computer Science and Technology, Nanjing University of Aeronautics and Astronautics, Nanjing 211106, China (e-mail: \{xumengting, dqzhang\}@nuaa.edu.cn)}
\thanks{Daoqiang Zhang is the corresponding author}
\thanks{Manuscript received May 09, 2021; revised May 09, 2021 .}}

%
%

\markboth{Journal of \LaTeX\ Class Files,~Vol.~14, No.~8, August~2015}%
{Shell \MakeLowercase{\textit{et al.}}: Bare Demo of IEEEtran.cls for IEEE Journals}
%



\maketitle

\begin{abstract}
Adversarial training has shown excellent high performance in defensing against adversarial examples. Recent studies demonstrate that examples are not equally important to the final robustness of models during adversarial training, i.e., the so-called \textit{hard} examples which can be attacked easily exhibit more influence than robust examples on the final robustness. Therefore, guaranting the robustness of hard examples is crucial for improving the final robustness of the model.
However, defining an effective heuristics to search hard examples is still difficult.
In this paper, inspired by information bottleneck (IB) principle, we uncover that example with high mutual information of the input and its associated latent representation is more likely to be attacked. Based on this observation, we propose a novel and effective adversarial training method (InfoAT). InfoAT is encouraged to find examples with high mutual information and exploit them efficiently to improve the final robustness of models.
Experimental results show that InfoAT achieves the best robustness among different datasets and models in comparison with several state-of-the-art methods.
\end{abstract}


%
\IEEEpeerreviewmaketitle

\section{Introduction}
%
%
%
%
\IEEEPARstart{D}{espite} the great success of deep neural networks on diverse tasks such as image classification~\cite{krizhevsky2012imagenet}, speech recognition and natural language processing~\cite{taigman2014deepface}. 
Recent studies have shown that their adversarial robustness is very poor~\cite{szegedy2014intriguing,goodfellow2014explaining}. When attacked by adversarial examples which added visually imperceptible perturbations, deep neural networks can be easily misleaded by a high confidence. 
The vulnerability of deep neural networks to adversarial examples may pose a huge threat to applications and services that are highly dependent on security performance, such as autonomous driving systems~\cite{levinson2011towards,chen2015deepdriving}, computer-aided disease diagnosis based on medical images~\cite{xu2021towards}, and malware detection protocols~\cite{marpaung2012survey}.

Considering the significance of adversarial robustness in neural networks, a range of defense methods have been proposed~\cite{guo2018countering,meng2017magnet,buckman2018thermometer,zou2021adversarial,xu2021towards}. 
Sun et al.~\cite{sun2019adversarial} achieved high level of robustness by projecting both clean and adversarially attacked input images into a low-dimensional quasi-natural image space. Zhang et al.~\cite{zhang2021robust}  proposed to progressively minimize the divergence of different distributions, such that the gap between the clean and adversarial domains in the intermediate feature space and decision space is minimum.
\begin{figure}[t]
	\centering      
	\subfigure[]{
		\label{fig:mu-info}
		\begin{minipage}{0.8\columnwidth}
			\centering                                                          
			\includegraphics[width=1\columnwidth]{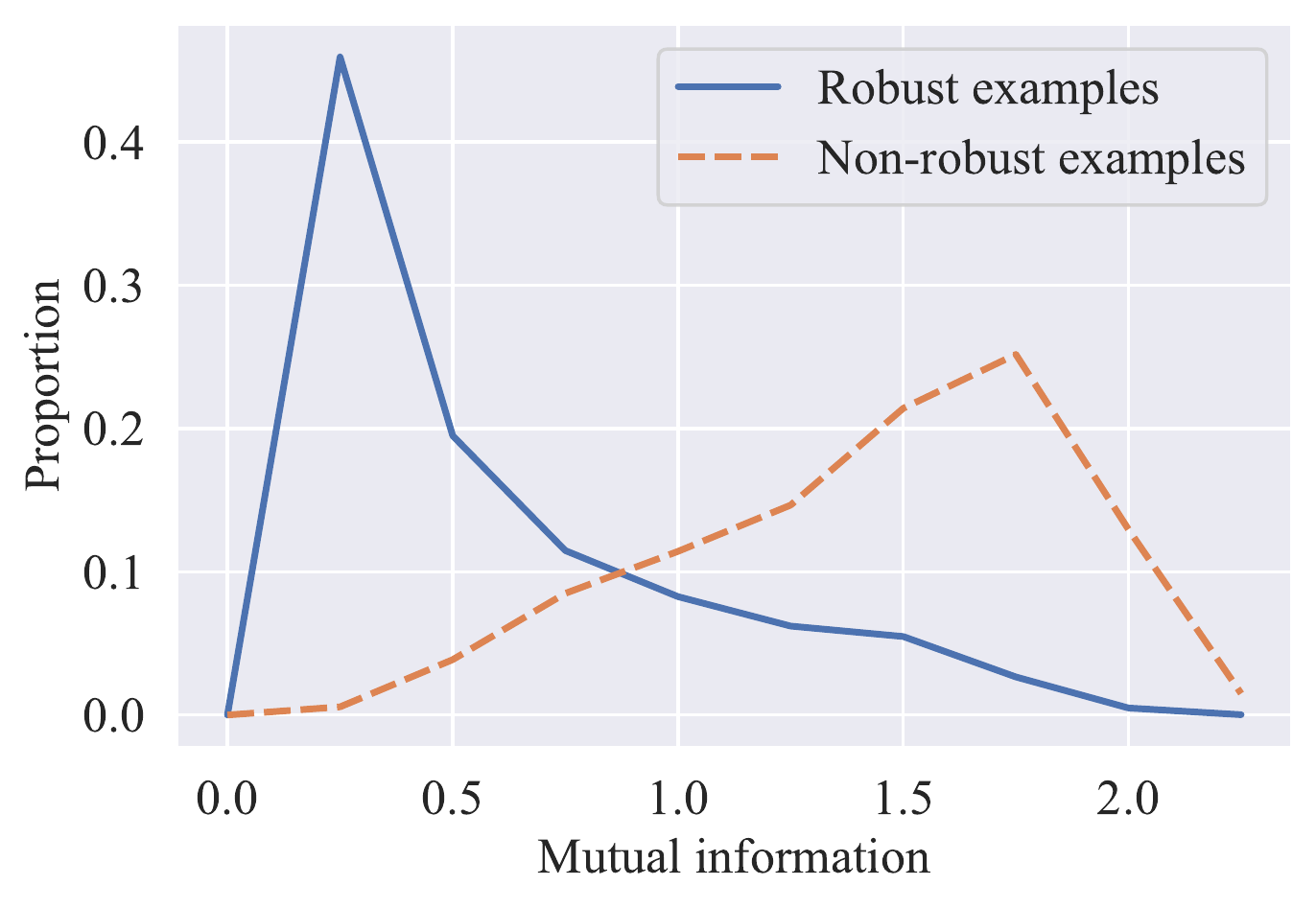}           
	\end{minipage}}
	\subfigure[]{ 
		\label{fig:mu-info-unrobust}                  
		\begin{minipage}{0.8\columnwidth}
			\centering                                                          
			\includegraphics[width=1\columnwidth]{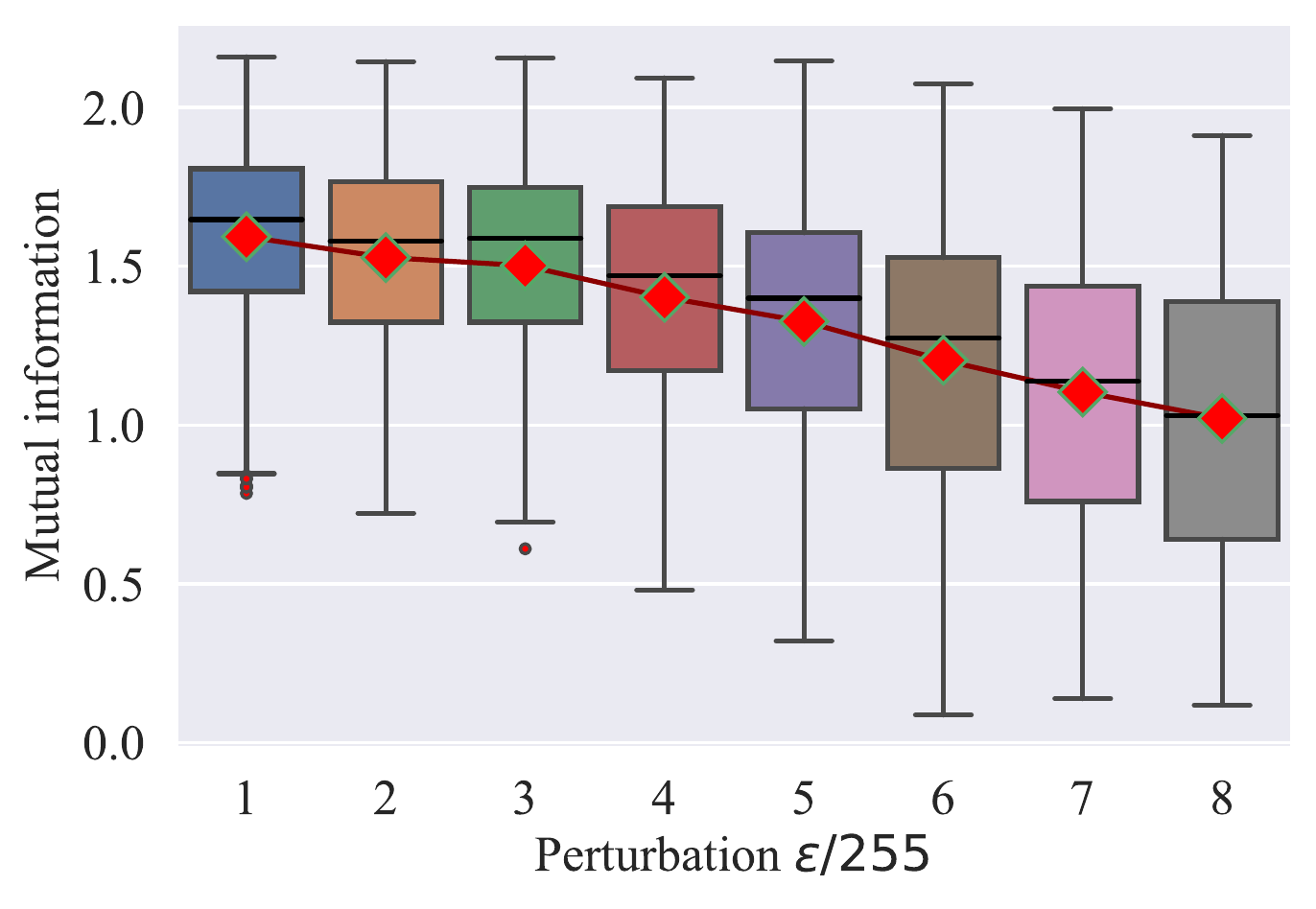}           
	\end{minipage}}   
	\caption{Examples with high mutual information are more likely to be attacked. (a) Most non-robust (can be successfully attacked) examples have higher mutual information (dashed orange line) than robust examples (blue line). (b) The minimum adversarial perturbation required for examples with different mutual information to be successfully attacked. The larger the example mutual information, the smaller the perturbation required for a successful attack.  The dataset is CIFAR-10. The model is ResNet-18 trained by TRADES~\cite{zhang2019theoretically}. The attack method is Project Gradient Descent (PGD)~\cite{madry2018towards} with 10 steps. }         
	\label{fig: motivation}                                             
\end{figure}
Among them, adversarial training, which continuously adds newly generated adversarial examples in the neural network training process, has been proven to be one of the most effective defense strategies against adversarial attacks.
Many adversarial training strategies employ the ``worst-case" adversarial example to maximize the loss for updating the current model such as standard adversarial training (AT)~\cite{madry2018towards}, TRADES~\cite{zhang2019theoretically}, and robust self-training (RST)~\cite{carmon2019unlabeled}.
Zhang et al.~\cite{zhang2019defense} generated adversarial images for adversarial training through feature scattering in the latent space, which is unsupervised in nature and avoids label leaking.
Liu et al. designed a GanDef, which utilizes a competition game to regulate the feature selection during the adversarial training~\cite{liu2019gandef}~\cite{liu2019zk}. And \cite{park2020effectiveness} showed that adversarially training a network with an attack method helps defending against that particular attack method, but has limited effect for other attack methods.
Furthermore, recent studies~\cite{wang2019improving,zhang2020geometry} have shown examples along with their adversarial variants are not equally important to the robustness of models during adversarial training. 
Specifically, the so-called \textit{hard} examples (can be attacked easily) exhibit more influence than robust examples on the final robustness.
Therefore, it is crucial to define an effective heuristics to search hard examples for improving the robustness of model.

In this paper,
we propose a novel and effective adversarial training strategy (InfoAT) from an information theoretic perspective. It uses the Information Bottleneck (IB) principle~\cite{tishby2000information} to find the hard examples. 
Recently, the studies of IB principle to deep neural network training have developed rapidly~\cite{achille2018information,amjad2019learning}, but its effect on the effectiveness of adversarial training still remains unclear.
The IB principle shows that when the valid information of the input cannot be well compressed by the model, the neural network cannot generalize the out-of-distribution data well, that is, the mutual information between the input data and its related latent representation is high.
As shown in Fig.~\ref{fig:mu-info}, hard (i.e., non-robust, can successfully attacked by Project Gradient Descent (PGD)~\cite{madry2018towards}  attack) examples are more likely to have higher mutual information than robust examples. What's more, Fig.~\ref{fig:mu-info-unrobust} shows the larger the mutual information of example, the smaller the perturbation required for a successful attack. 
These phenomena show that example with high mutual information is more vulnerable than other examples. This is because when the mutual information is high, the latent representation of the network has learned a large number of features of the input image.
And these features often contain some redundant information and non-robust features, so when there is adversarial perturbations in the image, the model will be easily fooled.
Motivated by these conceptual and experimental observations, we adopt the mutual information of the original examples between the input and latent distribution with the current model to select the hard examples, and guarantee them to be robust against attackers to improve the robustness of the model.

Our main contributions are:
\begin{itemize}
	\item We show that the example with high mutual information is more likely to be attacked in Fig.~\ref{fig: motivation}. 
	High mutual information means that the latent representation of the network contains a lot of information about the input image, including the redundant information and adversarial perturbations that are useless for the prediction results, which makes the network more vulnerable on these examples.
	\item We propose a novel and effective adversarial training (InfoAT) using the IB principle, and address the problem of selecting hard example with high mutual information. To the best of our knowledge, it has not been studied yet.
	\item We show the effectiveness of InfoAT by different networks and datasets. Specifically, 
	under the most common comparison setting (PGD attack with 20 steps on CIFAR- 10, with  $L_\infty$ perturbation $\epsilon=8/255$), InfoAT achieves the state-of-the-art robustness of 60.46\% with WideResNet, which improves $\sim 7\%$ over standard AT~\cite{madry2018towards}, $\sim 6\%$ over GanDef~\cite{liu2019gandef}, $\sim4\%$ over TRADES~\cite{zhang2019theoretically} and $\sim2\%$ even over MART~\cite{wang2019improving}. 
\end{itemize}

\section{Background and Related Work}
\subsection{Base Classifier}
For a $K$-class ($K\geq 2$) classification problem, denote a dataset $\{(\bm{x}_i ,y_i)\}_{i=1,\cdots,n}$ with distribution $\bm{x}_i \in \mathbb{R}^d $ as natural input and $y_i \in \{1,\cdots,K\}$ represents its corresponding true label, a classifier $h_{\bm{\theta}}$
with parameter $\bm{\theta}$ predicts the class label of an input example $\bm{x}_i$:
\begin{equation}
h_{\bm{\theta}}(\bm{x}_i)=\mathop {\arg\max}_{k=1,\cdots,K} \bm{p}_k(\bm{x}_i,\bm{\theta}), 
\end{equation}
where
\begin{equation}
\bm{p}_k(\bm{x}_i,\bm{\theta})=\exp(\bm{z}_k(\bm{x}_i,\bm{\theta}))/ \sum_{k^\prime=1}^K \exp(\bm{z}_{k^\prime}(\bm{x}_i,\bm{\theta})),
\end{equation}
$\bm{z}_k(\bm{x}_i,\bm{\theta})$ is the network logits output with respect to class $k$, and $\bm{p}_k(\bm{x}_i,\bm{\theta})$ is the probability (softmax on logits) that $\bm{x}_i$ belonging to class $k$.

\subsection{Adversarial Training}
One of the most successful empirical defenses to date is adversarial training. It was first proposed in~\cite{szegedy2014intriguing} and~\cite{goodfellow2014explaining}, where they showed that adding adversarial examples to the training set can improve the robustness against attacks. More recently, \cite{madry2018towards} formulated adversarial training as a min-max optimization problem and demonstrated that adversarial training with Project Gradient Descent (PGD) attack leads to empirical robust model. Specifically, the objective function of \textit{standard adversarial training} (AT) is:
\begin{equation}
\label{equ:AT}
\mathcal{L}_{\rm AT}=\min \frac{1}{n} \sum_{i=1}^{n} \ell (\htheta(\hat \bx_i^{\prime}) ,y_i),
\end{equation}
where
\begin{equation}
\label{equ:ATX}
\hat \bx_i^{\prime} = \arg \max_{\bx_i^{\prime}\in \mathcal{B}_\epsilon(\bx_i)} \ell (\htheta(\bx_i^\prime),y_i),
\end{equation}
$\ell$($\cdot$) represents the loss function such as commonly used cross entropy loss, and $\mathcal{B}_\epsilon(\bm{x}_i)=\{\bm{x}_i^\prime:||\bm{x}_i^\prime-\bm{x}_i||_p\leq\epsilon\}$ denotes the $L_p$-norm ball centered at $\bm{x}_i$ with radius $\epsilon$. In this work, we only focus on the $L_\infty$-norm.

Based on the primary AT framework, the recent researches have proposed many improvements from different perspectives, and demonstrated strong performance.
TRADES~\cite{zhang2019theoretically} trades adversarial robustness off against accuracy. 
The objective function of TRADES is a linear combination of natural loss and regularization term. Its regularization term uses Kullback-Leibler (KL) divergence to constrain the output probability between the original example and its adversarial example.
The objective function of TRADES can be described as follows:
\begin{equation}
\begin{split}
\mathcal{L}_{\rm TRADES} = & \min \frac{1}{n} \sum_{i=1}^{n} \{\ell (\htheta(\bx_i),y_i)\\&+\lambda \cdot  \mathcal{\rm KL}(\bp(\bx_i,\btheta)||\bp(\hat \bx_i^\prime,\btheta))\},
\end{split}
\end{equation}
where
\begin{equation}
\hat \bx_i^{\prime} = \arg \max_{\bx_i^{\prime}\in \mathcal{B}_\epsilon(\bx_i)} \mathcal{\rm KL}(\bm{p}(\bm{x}_i,\btheta)||\bm{p}(\bx_i^\prime,\btheta)),
\end{equation}
\begin{equation}
\mathcal{\rm KL}(\bm{p}(\bm{x}_i,\btheta)||\bm{p}(\hat \bx_i^\prime,\btheta))=\sum_{k=1}^K \bm{p}_k(\bm{x}_i,\theta)\log \frac{\bm{p}_k(\bm{x}_i,\theta)}{\bm{p}_k(\hat\bx_i^\prime,\theta)}
\end{equation}
measures the difference of two distributions. 
Furthermore, 
MART~\cite{wang2019improving} gives the misclassified examples and correctly classified examples different weights during adversarial training and adopts a regularized adversarial loss involving both adversarial and natural examples.
The objective function of MART is as follows:
\begin{equation}
\begin{split}
&\mathcal{L}_{\rm MART}=\min \frac{1}{n}\sum_{i=1}^{n}\{{\rm BCE} (\htheta(\hat\bx_i^\prime),y_i) \\&+\lambda \cdot {\rm KL}(\bp(\bx_i,\btheta)||\bp(\hat\bx_i^\prime,\btheta)) \cdot (1-\bp_{y_i}(\bx_i,\btheta))\},
\end{split}
\end{equation}
where
\begin{equation}
\label{equ:MARTX}
\hat \bx_i^{\prime} = \arg \max_{\bx_i^{\prime}\in \mathcal{B}_\epsilon(\bx_i)} \ell (\htheta(\bx_i^\prime),y_i),
\end{equation}
where ${\rm BCE(\cdot)}$ is boosted cross-entropy loss proposed by~\cite{wang2019improving}, 
and $\bp_{y_i}(\bx_i,\btheta)$ is the probability of $\bx_i$ belonging to class $y_i$.

\subsection{Information Bottleneck (IB) Principle}
Different from these methods, our method selects hard examples to efficiently improve the model robustness using the information bottleneck principle.
The Information bottleneck (IB)~\cite{tishby2000information} is a principled method of finding a latent representation $Z$, which represents the information contained in the input variable $X$ about the output $Y$.
The mutual information of $X$ and $Z$ can be written as $I(X;Z)$. 
$I(X;Z) ={\rm KL}(P_{XZ}||P_{X}P_{Z})$, where $P_{XZ}$ is the joint distribution and $P_{X}, P_{Z}$ are the marginal distributions of input variables $X,Z$, respectively.
Intuitively, $I(X;Z)$ measures the uncertainty of $X$ given $Z$. It not only reflects how much $Z$ compresses $X$ through $I(X;Z)$, but also reflects how well $Z$ predicts $Y$ through $I(Y;Z)$. In practice, this IB principle is achieved by minimizing the following IB Lagrangian:
\begin{equation}
\label{equ:IB}
\min \{I(X;Z)-\lambda \cdot I(Y;Z)\},
\end{equation}
where $\lambda$ represents a positive parameter, used to controls the trade-off between compression and prediction. 
Through the compression item $I(X;Z)$ to control the amount of compression in $Z$, we can adjust the required features of the training model, such as  data augmentation~\cite{zhao2020maximum}, generalization error~\cite{amjad2019learning}, and detection of out-of-distribution data~\cite{alemi2018uncertainty}. 
In this paper, we use IB principle to select hard examples with high mutual information during adversarial training.
Specifically, we use $I(X=\bx_i;Z=\bz_i)$ to denote the point-wise mutual information~\cite{ince2017measuring} of the input $\bx_i$ and its latent representation $\bz_i$, which means the relevant information that $\bz_i$ provides about $\bx_i$~\cite{tishby2000information}. Hereafter, to avoid notational clutter, we will shorten $``X=\bx_i"$ to $\bx_i$, $``Z=\bz_i"$ to $\bz_i$ and $``Y=y_i"$ to $y_i$ in all conditionals and point-wise mutual information theoretic functions.

\section{Methods}

\subsection{Regularizations Using IB Principle}
Note that the adversarial risk in Equations~(\ref{equ:AT})-(\ref{equ:ATX}) treat all examples equally,
regardless of the different effects of these examples on the robustness of the model.
Therefore, the main idea of our paper is to apply the IB principle to adversarial training to exploit hard examples efficiently so as to improve the robustness of models.

According to Equation~(\ref{equ:IB}), 
when the natural input information cannot be well compressed by the model, the neural network cannot well generalize the out-of-distribution data, that is, the mutual information of the natural input and its related latent representation is very high,
as confirmed in Fig.~\ref{fig: motivation}.  Based on these observations, we consider adding the IB-weighted regularization of natural input to the standard cross-entropy loss. 
The IB-weighted regularization on natural input is defined as follows:
\begin{equation}
\mathcal{L}_{\rm IB}(\bx,\btheta) =\frac{1}{n} \sum_{i=1}^{n}I(\bx_i;\bz_i) \cdot \mathds{1}(\htheta(\hat\bx_i^\prime) \neq \htheta(\bx_i)),
\end{equation}
where $\hat\bx_i^\prime$ is the ``worst-case" adversarial examples with the $\epsilon$-ball of natural input $\bx_i$, $I(\bx_i;\bz_i)$ is the point-wise mutual information of the input $\bx_i$ and its latent representation $\bz_i$, which was used to select the hard examples (this will be large for hard examples and small for robust examples). $\mathds{1}(\htheta(\hat\bx^\prime) \neq \htheta(\bx))$ aims to encourage the output of neural network to be stable against example which has high mutual information.
$\mathds{1}(\htheta(\hat\bx_i^\prime) \neq \htheta(\bx_i))=1$ when $\htheta(\hat\bx_i^\prime) \neq \htheta(\bx_i)$, and $\mathds{1}(\htheta(\hat\bx_i^\prime) \neq \htheta(\bx_i))=0$ otherwise.

Moreover, \cite{ilyas2019adversarial} disentangles adversarial examples as a natural consequence of non-robust features. A common belief is that adversarial robustness comes from feature representations learned through adversarial examples~\cite{santurkar2019image,ilyas2019adversarial}. From this perspective, the network is expected to maintain more information of adversarial examples (robust features), i.e., the mutual information of the adversarial input and its associated latent adversarial representation should to be high. Therefore, the newly added adversarial regularization on adversarial examples is defined as follows:
\begin{equation}
\label{equ:reg}
\mathcal{L}_{\rm Reg}(\hat\bx^\prime,\btheta) = \frac{1}{n}\sum_{i=1}^{n}-I(\hat\bx_i^\prime;\hat\bz_i^\prime),
\end{equation}
where $I(\hat \bx_i^\prime;\hat\bz_i^\prime)$ is the point-wise mutual information of the adversarial example $\hat \bx_i^\prime$ and its latent representation $\hat\bz_i^\prime$.

Finally, by combining the proposed IB-weighted regularization and adversarial regularization in the adversarial training framework, we can train a network by proposed InfoAT that minimizes the following risk:
\begin{equation}
\begin{split}
\mathcal{L}_{\rm min}(\bx,\btheta)& = \min_\btheta \frac{1}{n} \sum_{i=1}^{n} \{\ell (\htheta(\hat \bx_i^\prime),y_i)\\&+\lambda \cdot I(\bx_i;\bz_i) \cdot \mathds{1}(\htheta(\hat \bx_i^\prime) \neq \htheta(\bx_i))  \\&-\beta \cdot I(\hat \bx_i^\prime;\hat\bz_i^\prime) \},
\end{split}
\end{equation}
where $\lambda,\beta$ are tunable scaling parameters which are fixed for all training examples, $\ell(\cdot)$ is the standard cross-entropy loss. Next, we will describe how to generate the ``worst-case" adversarial example $\hat \bx_i^\prime$ in our InfoAT.

\begin{algorithm}[t]
	\caption{Adversarial training using the information bottleneck principle (InfoAT)}
	\label{alg:algorithm}
	\textbf{Input}: $\{\bx_i,y_i\}_{i=1\cdots n}$: training data; $T_O$: outer iteration number; $T_I$: inner iteration step; $\epsilon$: maximum perturbation; $\alpha_I$: step size for inner optimization; $\alpha_O$: step size for outer optimization; $m$: the number of each batch size; $\lambda$ and $\beta$: tunable scaling parameter.\\
	\textbf{Initialization}: Standard random initialization of $\htheta$ \\
	\textbf{Output}: Robust classifier $\htheta$
	\begin{algorithmic}[1] 
		\FOR {$t=1,\cdots T_O$}
		\STATE Uniformly sample a minibatch of training data $B^{(t)}$
		\FOR{$\bx_i\in B^{(t)}$}
		\STATE $\bx_i^\prime=\bx_i+\epsilon\cdot \xi$, with $\xi \sim \mathcal{U}(-1,1)$
		\STATE \# $\mathcal{U}$ is a uniform distribution
		\FOR{$s=1,\cdots,T_I$}
		\STATE $\bx_i^\prime \leftarrow \Pi_{\mathcal{B}_\epsilon(\bx_i)}(\bx_i^\prime+\alpha_I\cdot {\rm sign}(\nabla_{\bx_i^\prime} {\rm CE}(\bp(\bx_i^\prime),y_i)+\lambda \nabla_{\bx_i^\prime}H(\bp(\bx_i)) \cdot ||\bp(\bx_i^\prime) -\bp(\bx_i)||_2^2 ))$
		\STATE \# $\Pi(\cdot)$ is the projection operator
		\ENDFOR
		\STATE $\hat \bx_i^\prime \leftarrow \bx_i^\prime$
		\ENDFOR
		\STATE $\btheta \leftarrow \btheta-\alpha_O\frac{1}{m}\sum_{\bx_i\in B^{(t)}} \nabla_\btheta\{ -\beta \cdot H(\bp(\hat \bx_i^\prime)) +{\rm CE} (\bp(\hat \bx_i^\prime),y_i) +\lambda \cdot H(\bp(\bx_i)) \cdot ||\bp(\hat \bx_i^\prime) -\bp(\bx_i)||_2^2
		\}$
		\ENDFOR
	\end{algorithmic}
\end{algorithm}

Standard adversarial training essentially minimizes a lower bound on the inner maximum ``worst-case" loss, which will mislead an optimizer to search for a local optimal value~\cite{zhang2019towards}. To alleviate this pessimism, we also consider to incorporate IB regularization into maximization phase. 
In maximization phase, the model parameters $\btheta$ are fixed, and $\bz_i$ is a deterministic representation of input $\bx_i$, i.e., any given input is mapped to a fixed representation. Consequently, it holds that $I(\bx_i,\bz_i)$ is a different constant for different input $\bx_i$.
Therefore, the IB regularization can be used to search more out-of-distribution adversarial data (i.e., the mutual information of its natural example is high) in the maximization phase. 
From this perspective, we extend the proposed IB regularization to maximization phase to search for the ``worst-case" (i.e., more out-of-distribution) data to improve model robustness. 
We have verified its effectiveness in experiments. 
The generation of $\hat \bx_i^\prime$ can be defined as:

\begin{equation}
\label{equ:advexp}
\begin{split}
\hat \bx_i^{\prime} &= \arg \max_{\bx_i^{\prime}\in \mathcal{B}_\epsilon(\bx_i)} \{\ell (\htheta(\bx_i^\prime),y_i) \\&+\lambda \cdot I(\bx_i;\bz_i) \cdot \mathds{1}(\htheta(\bx_i^\prime) \neq \htheta(\bx_i))\}.
\end{split}
\end{equation}

\subsection{Optimization for InfoAT}
As analysed in previous section, we derived the adversarial risk of InfoAT based on point-wise mutual information (i.e., $I(\bx_i;\bz_i)$, $I(\hat\bx_i^\prime;\hat\bz_i^\prime)$) and 0-1 loss (i.e., $\mathds{1}(\htheta(\bx_i^\prime) \neq \htheta(\bx_i)))$. However, mutual information in example-level is shown to be intractable in the literature~\cite{belghazi2018mutual,song2019understanding} since $\bz_i$ is high dimensional and optimization over 0-1 loss is also difficult, therefore directly optimizing this objective is challenging. We then propose the optimization strategies, by surrogating mutual information with entropy, and replacing 0-1 loss with proper distance function, which are both physically meaningful and computationally tractable.

\subsubsection{Surrogating Mutual Information with Entropy}
A variety of methods have been proposed for estimating mutual information~\cite{kandasamy2015nonparametric,moon2017ensemble,walters2009estimation}. One of the most effective estimators is the mutual information neural estimator (MINE)~\cite{belghazi2018mutual}. It proves that, for all $\eta \ge 0$, there exists $M \in \mathbb{Z}$ that satisfies certain constraints, such that for any $m \geq M$, $|\hat{I}_m(X;Z)-I(X;Z)|\leq \eta$ almost holds, where $m$ is the number of sampling examples. 

However, our work exploits mutual information of each example (instead of on $m\geq M$ examples) to select the hard example. From this perspective, MINE may not be applicable in our case. 
On example-level, the mutual information $I(\bx_i,\bz_i)$ for example $\bx_i$ is still intractable.
Our approach in this paper is that, when in the task of classification scenarios, $I(\bx_i;\bz_i)$ can be effectively approximated during training. As we will show, this process can be effectively realized by the entropy $H(\tilde y_i)$ predicted by the network, which is a tractable lower bound of $I(\bx_i;\bz_i)$,
$H(\cdot)$ is the Shannon entropy and $\tilde{y_i}$ is the output prediction of current model $\htheta(\cdot)$.

As described in ~\cite{amjad2019learning}, a deep neural network can be viewed as a Markov chain of continuous representation of the input, where the information flow follows the following structure: $x_i \rightarrow z_i \rightarrow \tilde{y}_i$, where $\tilde{y}_i$ is network prediction.
According to the data processing inequality~\cite{cover1999elements}, $I(x_i;z_i) \geq I(x_i;\tilde{y}_i)$ can be got, where $I(x_i; \tilde{y}_i) =H(\tilde{y}_i)-H(\tilde{y_i}|x_i)$. 
In other words, when calculating mutual information during the adversarial training process, the model parameter $\btheta$ is fixed, and any given input $x_i$ is mapped to a single class. That is, $H(\tilde{y_i}|x_i)=0$. After putting all these together, we have the following proposition.

\begin{proposition}
	Consider a deterministic neural network, the parameter $\btheta$ of which is fixed. Given an input $x_i$, a network prediction $\tilde{y}_i$, and a latent representation $z_i$. Then, the entropy $H(\tilde{y}_i)$ is the lower bound of the mutual information $I(x_i;z_i)$, i.e.,
	\begin{equation}
	I(x_i;z_i) \geq I(x_i;\tilde{y}_i) =H(\tilde{y}_i) -H(\tilde{y_i}|x_i) =H(\tilde{y}_i).
	\end{equation}
\end{proposition}

There are two important benefits of surrogating mutual information with entropy to be discussed. \textit{First}, $H(\tilde{y}_i)$ is the lower bound of $I(x_i;z_i)$, so it is the most representative term for choosing hard example which is non-robust to adversarial attack (i.e., a high entropy example indicates a high mutual information). \textit{Besides}, $H(\tilde{y}_i)$ can be easily computed from the softmax output of a classification network.
Therefore, for point-wise mutual information, we surrogate $I(\bx_i;\bz_i)$ with $H(\bp(\bx_i))$, $I(\hat\bx_i^\prime;\hat\bz_i^\prime)$ with $H(\bp(\hat \bx_i^\prime))$.

\begin{figure}[t]
	\centering      
	\subfigure[The choice of $\lambda$]{
		\label{fig:lambda}
		\begin{minipage}{0.48\columnwidth}
			\centering                                                          
			\includegraphics[width=1\columnwidth]{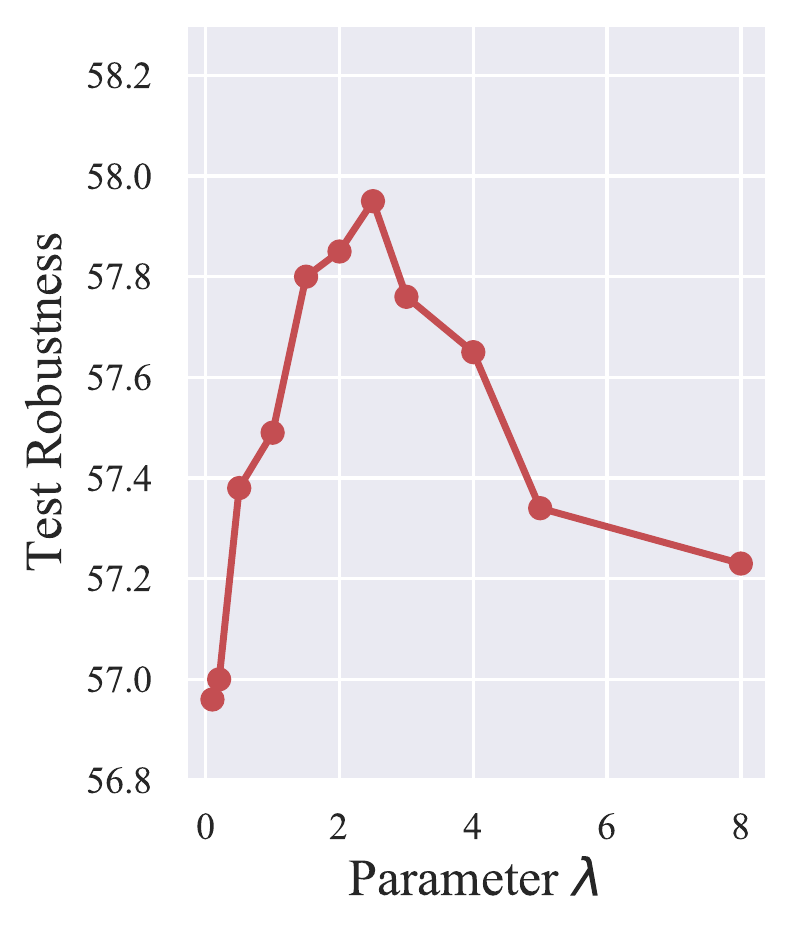}           
	\end{minipage}}
	\subfigure[The choice of $\beta$]{ 
		\label{fig:beta}                  
		\begin{minipage}{0.48\columnwidth}
			\centering                                                          
			\includegraphics[width=1\columnwidth]{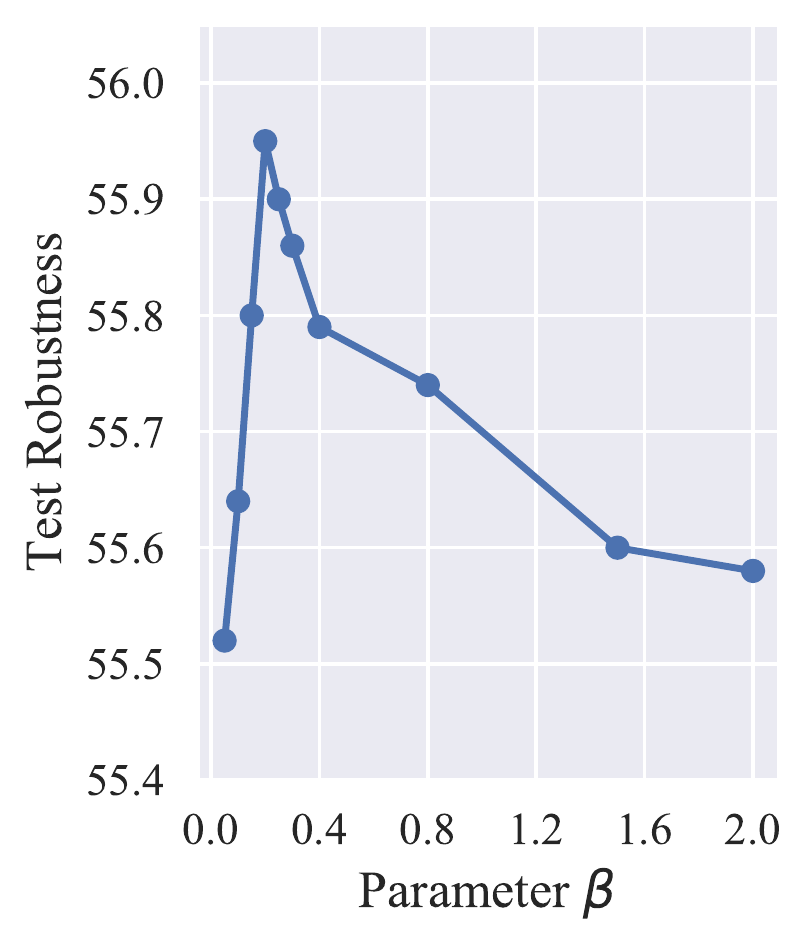}           
	\end{minipage}}     
	\caption{The sensitivities to parameters $\lambda$ and $\beta$. The units of ``Test robustness" is (\%). The dataset is CIFAR-10. The model is ResNet-18. The attack method is PGD with 20 steps.}                      
	\label{fig:ablation_app }                                             
\end{figure}

\begin{table}[t]
	\centering
	\caption{Robustness (\%) and accuracy (\%) based on MINE estimation mutual information on different parameter $\lambda$, where $\beta=0.2$.}
	\label{tab:app_mutual}
	\setlength{\tabcolsep}{5.0mm}{
		\begin{tabular}{c|c|cc}
			\toprule
			& $\lambda$ & Natural & $\rm PGD^{20}$ \\
			\midrule
			\multirow{5}{*}{layer \#1} & 0.5 & 84.42 & 56.71 \\
			& 1.0 & 84.53 & 56.79 \\
			& 1.5 & 84.69 & \textbf{57.21} \\
			& 2.0 & 82.52 & 56.09 \\
			\midrule
			\multirow{5}{*}{layer \#2} & 1.0 & 84.43 & 57.02 \\
			& 1.5 & 84.74 & 56.88 \\
			& 2.0 & 83.93 & 57.29 \\
			& 2.5 & 83.36 & \textbf{57.73} \\
			& 3.0 & 83.30 &57.46 \\
			\midrule
			\multirow{5}{*}{layer \#3} & 1.0 & 83.90 & 57.07 \\
			& 1.5 & 84.23 & 57.55 \\
			& 2.0 & 83.31 & 57.54 \\
			& 2.5 & 83.20 & 57.55 \\
			& 3.0 & 83.36 &\textbf{57.74} \\
			\midrule
			\multirow{5}{*}{layer \#4} & 1.0 & 84.15& 57.39 \\
			& 1.5 & 83.86 & \textbf{57.76} \\
			& 2.0 & 83.92 & 57.67 \\
			& 2.5 & 83.62 & 57.46 \\
			& 3.0 & 83.90 &57.40 \\
			\bottomrule
		\end{tabular}
	}
\end{table}

\begin{figure*}[t]
	\centering 
	\subfigure[Mutual information estimation]{
		\label{fig:mu}
		\begin{minipage}{1.0\columnwidth}
			\centering                                                          
			\includegraphics[width=1\columnwidth]{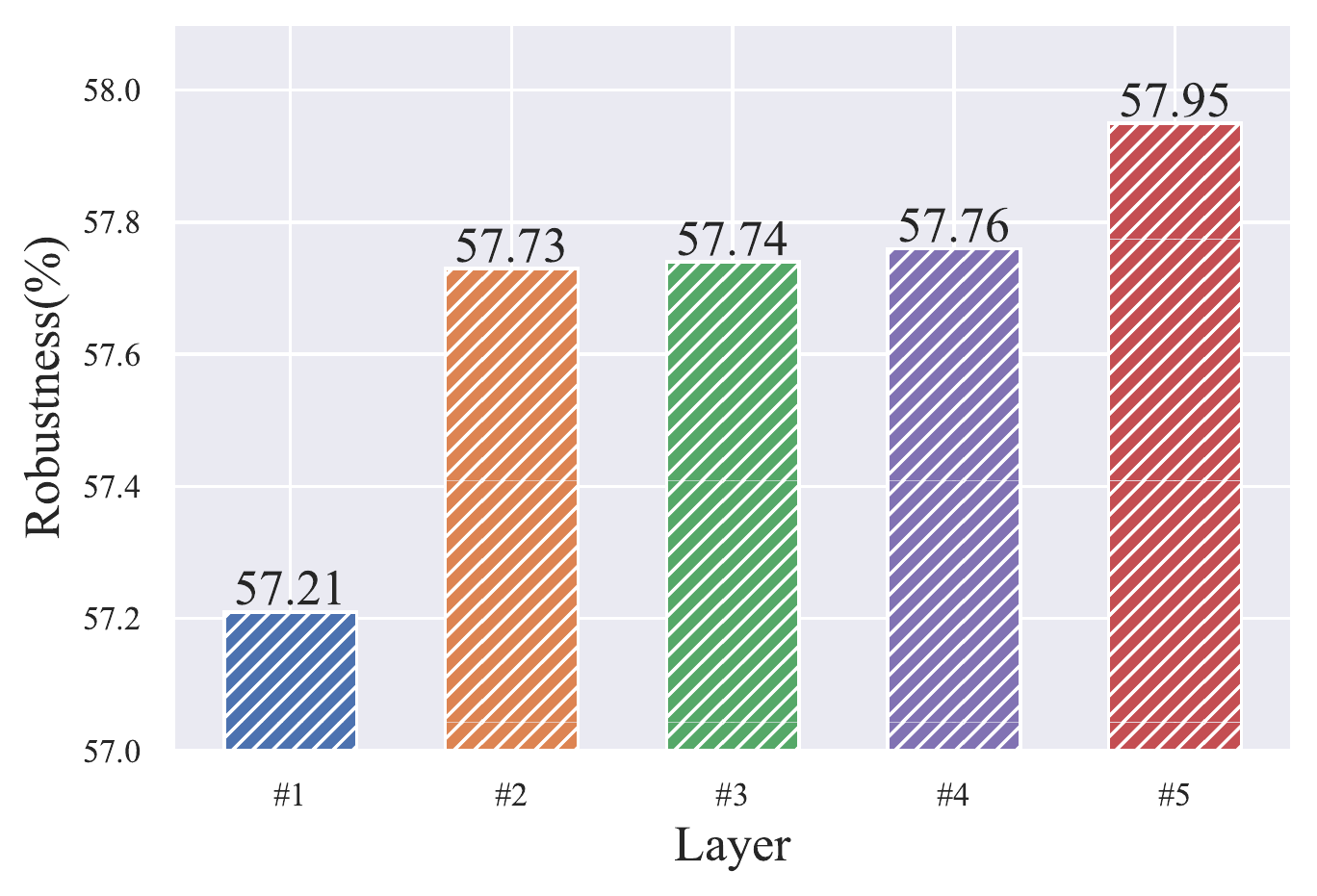}           
	\end{minipage}}     
	\subfigure[Different weighting methods]{
		\label{fig:weig}
		\begin{minipage}{1.0\columnwidth}
			\centering                                                          
			\includegraphics[width=1\columnwidth]{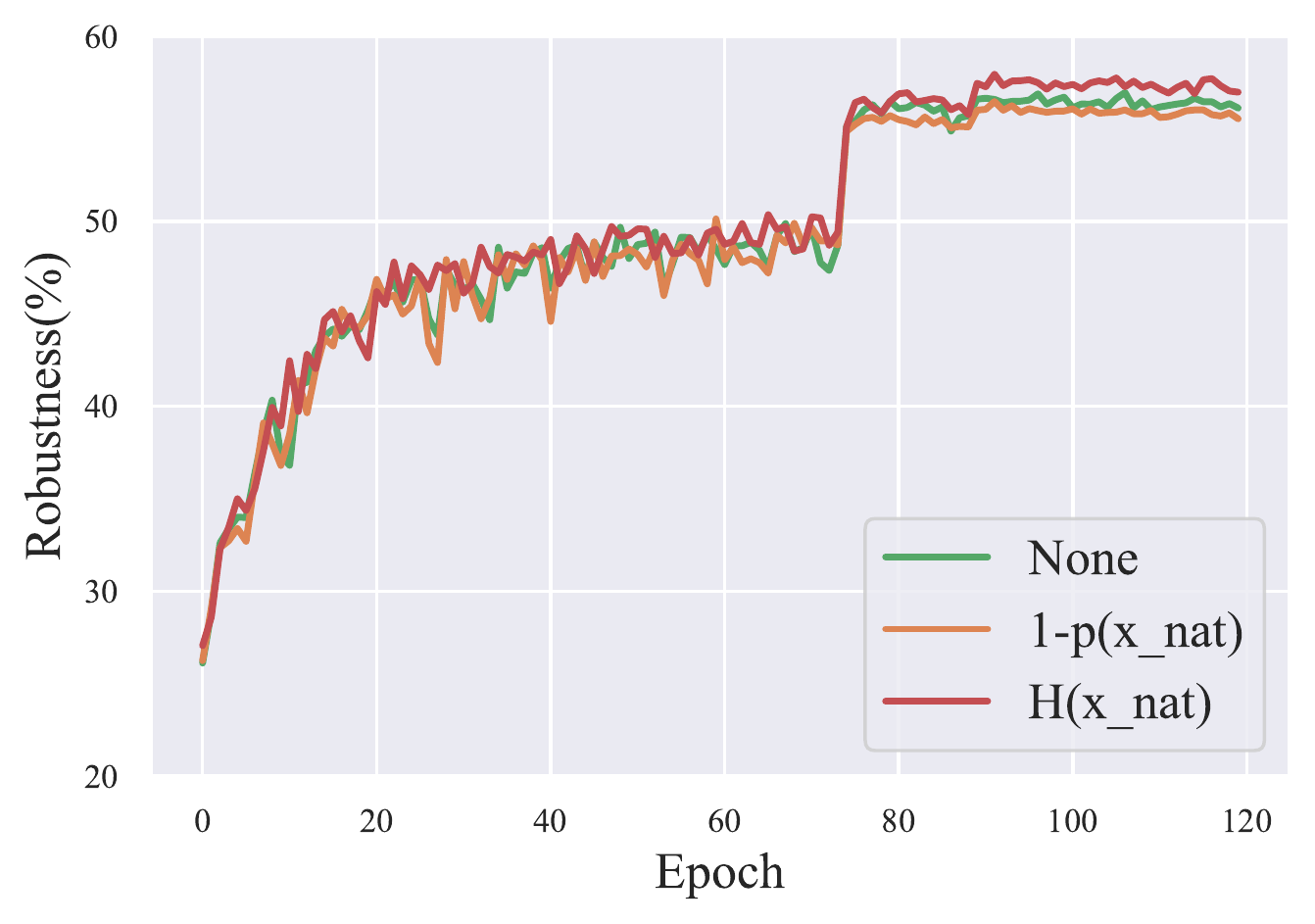}           
	\end{minipage}}
	\subfigure[Different divergences]{ 
		\label{fig:diver}                  
		\begin{minipage}{1.0\columnwidth}
			\centering                                                          
			\includegraphics[width=1\columnwidth]{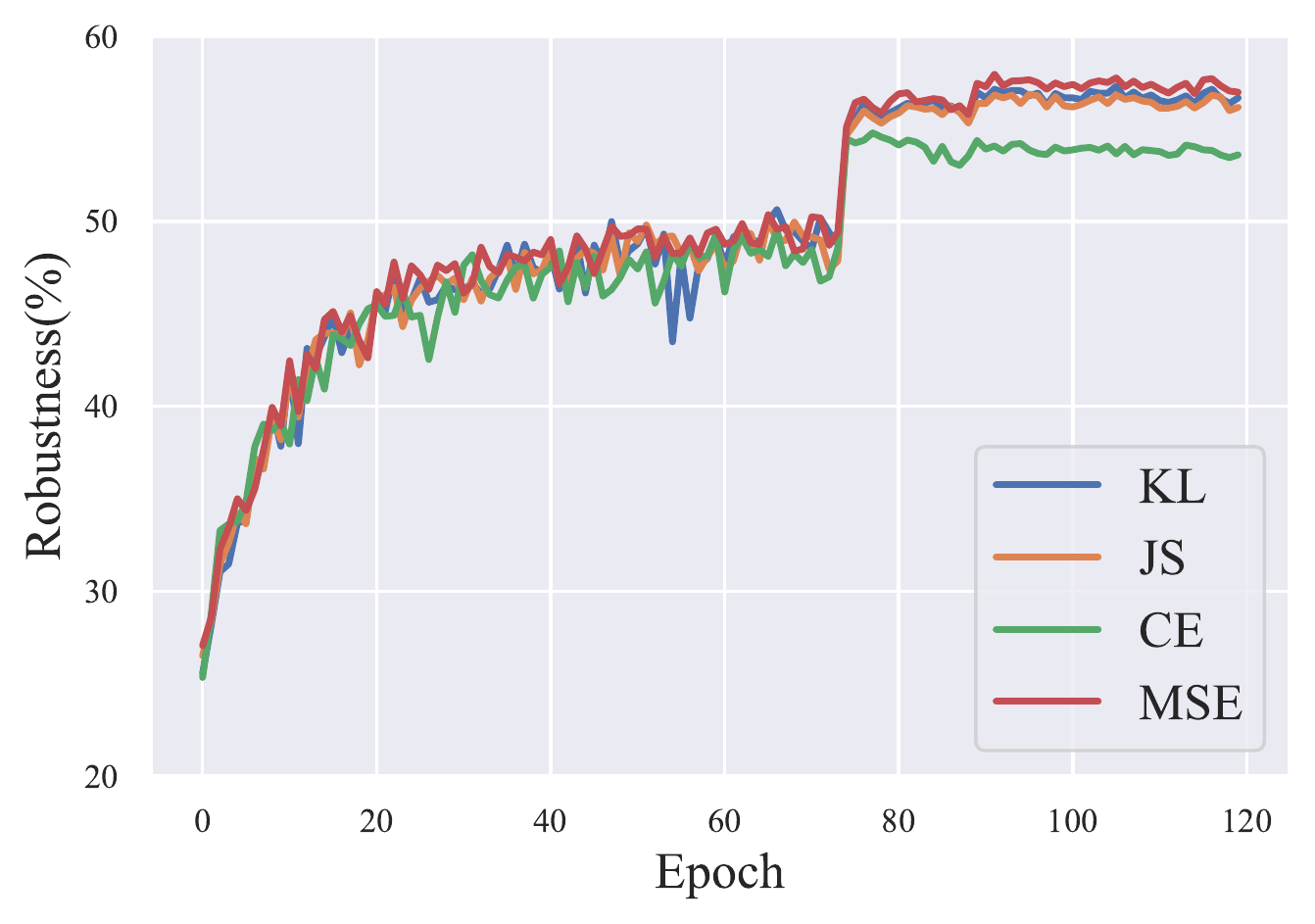}           
	\end{minipage}}
	\subfigure[Different regularizations]{  
		\label{fig:reg}                	  
		\begin{minipage}{1.0\columnwidth}
			\centering                                                          
			\includegraphics[width=1\columnwidth]{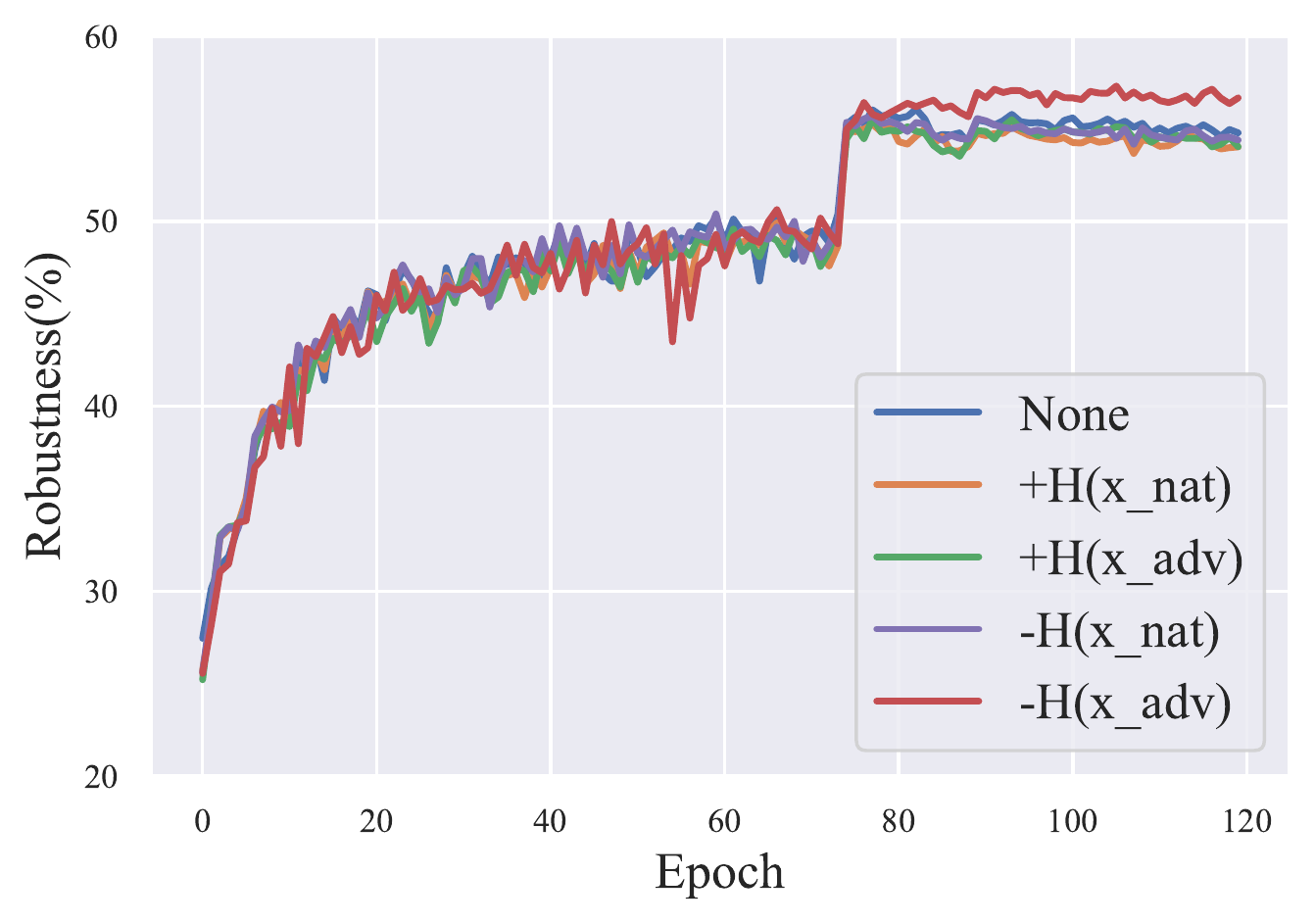}               
	\end{minipage}}     
	\caption{The comprehensive ablation experiments of InfoAT. In (a), $l \in \{\#1,\#2,\#3,\#4\}$ represent the mutual information estimated by MINE on different layer latent representations $\bz^{(l)}$, and $\#5$ represents the mutual information surragoted by entropy based on softmax output on the last layer $\#5$. In (b), (c), and (d), the red line represents our InfoAT method.
	The dataset is CIFAR-10. The model is ResNet-18. The attack method is PGD with 20 steps.}              
	\label{fig:ablation }                                             
\end{figure*}

\begin{table*}[t]
	\centering
	\caption{White-box robustness (accuracy (\%) on white-box test attacks, $\epsilon=8/255$) on CIFAR-10, CIFAR-100, and SVHN with the ResNet18.}
	\label{tab:white}
	\centering
	\begin{tabu}{c|c|ccccccccc|c}
		\toprule
		Dataset & Model & Natural & FGSM &$\rm PGD^{20}$ & $\rm PGD^{100}$ & $\rm PGD^+$ & $\rm CW_\infty^{20}$ & $\rm CW_\infty^{100}$ &BPDA& $\rm InfoPGD^{20}$&AA \\
		\midrule
		\multirow{6}{*}{CIFAR-10} &GanDef& \textbf{86.31}&63.63&49.25&44.71&45.09&49.44&45.62&48.18&47.57&47.13 \\
		&AT & 83.56 & 63.21& 51.47 & 49.54 & 49.71 & 51.23 & 49.51 & 52.91&52.29&47.25 \\
		& TRADES & 82.21 & 63.40 & 53.92 & 52.66 & 52.57 & 51.81 & 50.56 & 54.42&54.49&49.59 \\
		& MART & 79.24 & 63.74 & 56.01 & 54.79 & 54.62 & 51.12 & 49.76 & 56.07&56.63& 48.45 \\
		& MART+ & 79.52&64.19&56.69&55.22&55.10&51.37&50.01&55.67&56.17&48.50\\
		& InfoAT (Ours) & 83.23 & \textbf{67.23} & \textbf{57.95} & \textbf{56.15} & \textbf{56.14} & \textbf{52.91} & \textbf{51.17} & \textbf{56.84}&\textbf{57.08}& \textbf{49.67} \\
		\midrule
		\multirow{6}{*}{CIFAR-100} & GanDef& \textbf{65.15}&38.52&27.67&25.86&25.70&26.38&25.24&26.60&26.31&22.03 \\
		& AT& 57.28 &35.10 & 26.46 & 25.10 & 25.10 & 25.25 & 24.47 & 30.80&30.58& 22.89 \\
		& TRADES & 55.27 & 35.32 & 28.89 & 28.01 & 27.80 & 25.88 & 25.26 & 30.97&31.13& 24.56 \\
		& MART & 53.91 & 38.72 & 33.03 & 32.61 & 32.46 & 29.01 & 28.45 & 33.10&33.48& 26.67\\
		& MART+ & 47.84&36.96&32.30&31.93&31.82&27.00&26.62&31.94&32.73&26.50\\
		& InfoAT (Ours) & 58.23 & \textbf{40.62} & \textbf{34.08} &  \textbf{33.39} & \textbf{33.06} & \textbf{29.51} & \textbf{28.68}& \textbf{33.43}&\textbf{33.92}& \textbf{26.76} \\
		\midrule
		\textbf{\multirow{6}{*}{SVHN}} & GanDef& \textbf{93.25}&73.89&56.86&52.74&52.76&55.19&51.84&55.53&58.96&47.00\\
		& AT & 89.97 & 69.63& 54.78 & 51.61 & 51.76 & 50.87 & 48.78 & 53.53&53.02& 46.99  \\
		& TRADES & 89.26 & 71.43 & 58.01 & 55.87 & 55.51 & 54.77 & 52.97 &57.03&60.74& 50.82  \\
		& MART & 92.76 & 76.67 & 62.46 & 59.42 & 58.87 & 56.95 & 53.98 & 63.56&65.40& 49.25  \\
		& MART+ & 90.28&73.58&62.67&60.62&60.39&57.15&54.54&61.84&64.68&50.15\\
		& InfoAT (Ours) & 92.37 & \textbf{76.91} & \textbf{63.16} & \textbf{60.72} & \textbf{60.14} & \textbf{57.22} & \textbf{54.64} & \textbf{64.37}&\textbf{67.05}& \textbf{51.37}  \\
		\bottomrule 
	\end{tabu}
\end{table*}

\begin{table}[t]
	\centering
	\caption{Black-box robustness (accuracy (\%)) on CIFAR-10 with ResNet18 under SPSA attack with different batch sizes and $\epsilon=8/255$.}
	\label{tab:black}
	\centering
	\setlength{\tabcolsep}{1.1mm}{
		\begin{tabu}{c|cccc}
			\toprule
			Model & $\rm SPSA_{128}$ & $\rm SPSA_{256}$ & $\rm SPSA_{512}$ & $\rm SPSA_{1024}$ \\
			\midrule
			GanDef& 56.70&53.81&51.65&50.08\\
			AT & 58.23 & 55.77 & 53.90 & 52.74  \\
			TRADES & 58.69 & 56.48 & 54.82 & 53.60  \\
			MART & 57.41 & 55.50 & 53.84 & 52.52  \\
			MART+ & 57.73&55.62&53.20&52.58\\
			InfoAT (Ours) & \textbf{59.56} & \textbf{57.40} & \textbf{55.79} & \textbf{54.51}  \\
			\bottomrule
	\end{tabu}}
\end{table}

\begin{figure}[t]
	\centering
	\begin{minipage}{1\columnwidth}
		\centering                                                          
		\includegraphics[width=1.0\columnwidth]{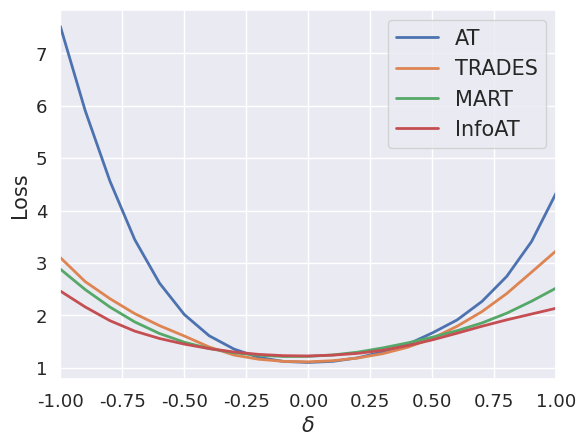}           
	\end{minipage}
	\caption{Weight loss surface across different methods on CIFAR-10 with ResNet18.}
	\label{fig:app_wloss}
\end{figure}

\subsubsection{Replacing 0-1 Loss with Distance Function}
For the 0-1 indicator function $\mathds{1}(\htheta(\bx_i^\prime) \neq \htheta(\bx_i))$, we use Mean Square Error (MSE) divergence as the surrogate loss function~\cite{kannan2018adversarial}, 
since $\htheta(\bx_i^\prime) \neq \htheta(\bx_i)$ represents that the output distributions of the adversarial example and the natural example are different. Thus, we have
\begin{equation}
{\rm{MSE}}(\bx_i^\prime, \bx_i) = ||\bp(\bx_i^\prime) -\bp(\bx_i)||_2^2.
\end{equation}
This will be large when the difference of distribution on adversarial examples and natural examples is high, and small otherwise.

\subsection{The Overall Objective Function}
Based on the above surrogate loss functions, we can illustrate the final objective function of InfoAT we proposed:
\begin{equation}
\label{equ:ob}
\begin{split}
\mathcal{L}_{\rm InfoAT}(\bx,\btheta) &=\min_{\btheta}\frac{1}{n}\sum_{i=1}^{n} \{
{\rm CE} (\bp(\hat \bx_i^\prime),y_i) \\ &+\lambda \cdot H(\bp(\bx_i)) \cdot ||\bp(\hat \bx_i^\prime) -\bp(\bx_i)||_2^2 \\&-\beta \cdot H(\bp(\hat \bx_i^\prime)) 
\},
\end{split}
\end{equation}
where
\begin{equation}
\begin{split}
\hat \bx_i^{\prime} &= \arg \max_{\bx_i^{\prime}\in \mathcal{B}_\epsilon(\bx_i)} \{{\rm CE} (\bp(\bx_i^\prime),y_i) +\\&\lambda \cdot H(\bp(\bx_i)) \cdot ||\bp(\bx_i^\prime) -\bp(\bx_i)||_2^2 \},
\end{split}
\label{equ:findadv}
\end{equation}
${\rm CE}(\cdot)$ is the cross-entropy loss. $\lambda,\beta$ are tunable scaling parameters used to balance three parts of the final loss, which are fixed for all training examples.
Algorithm~\ref{alg:algorithm} describes the complete training procedure of InfoAT.

\section{Experiments}
In this section, first we conduct a large number of ablation studies to fully understand our proposed InfoAT, and then we evaluate its robustness on the benchmark datasets under the black-box and white-box settings. Finally, we use the WideResNet to benchmark the state-of-the-art robustness and use unlabeled data for further robustness evaluation of our method.
\paragraph{Experimental Setup} Three commonly used datasets (i.e., CIFAR-10, CIFAR-100 and SVHN) and two networks (i.e., ResNet18 and WideResNet-34-10) are utilized in our experiments.
All the models with three datasets are trained using SGD, the momentum is 0.9, weight decay is $3.5 \times 10^{-3}$, and the initial learning rate is 0.01, which is divided by 10 at the 75-th, 90-th, and 100-th epoch. The attack used for training is ${\rm PGD}^{10}$ (PGD attack with 10 steps) with random start. The maximun perturbation $\epsilon=8/255$ and step size $\epsilon/4$, while the test attack is $\rm PGD^{10}$ with random start unless otherwise stated.

We perform all experiments on a desktop PC using a single GeForce RTX 2080 Ti GPU and 12-core Intel(R) Core(TM) i7-8700 CPU @ 3.20GHz. We implement training and certification in PyTorch.

\subsection{Ablation Study}
Here, we first investigate the parameter $\lambda \in [0,8]$ and $\beta \in [0.1,2.0]$ in InfoAT objective function. We present the results of $\lambda$ and $\beta$ in Fig.~\ref{fig:lambda} and Fig.~\ref{fig:beta}, respectively. InfoAT achieves good robustness across different choices of $\lambda$ and $\beta$. We choose $\lambda = 2.5, \beta=0.2$ for further experiments.

\begin{figure*}[t]
	\centering      
	\subfigure[]{
		\begin{minipage}{0.38\columnwidth}
			\centering                                                          
			\includegraphics[width=1\columnwidth]{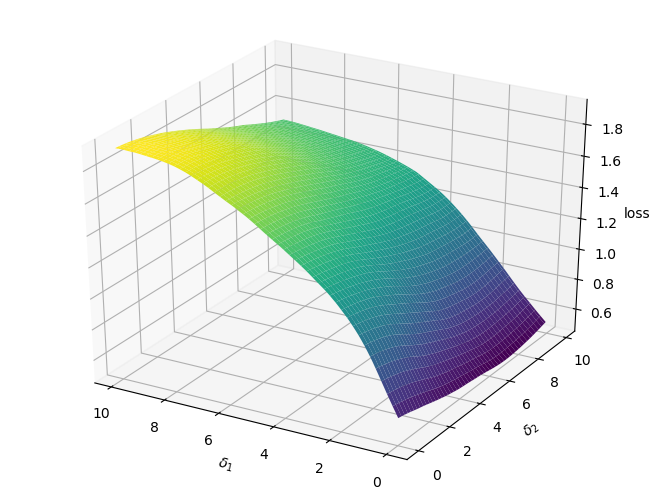}           
	\end{minipage}}
	\subfigure[]{
		\begin{minipage}{0.38\columnwidth}
			\centering                                                          
			\includegraphics[width=1\columnwidth]{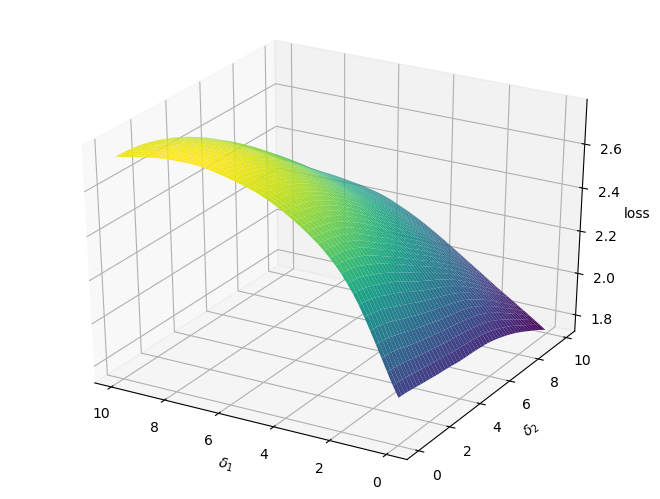}           
	\end{minipage}} 
	\subfigure[]{
		\begin{minipage}{0.39\columnwidth}
			\centering                                                          
			\includegraphics[width=1\columnwidth]{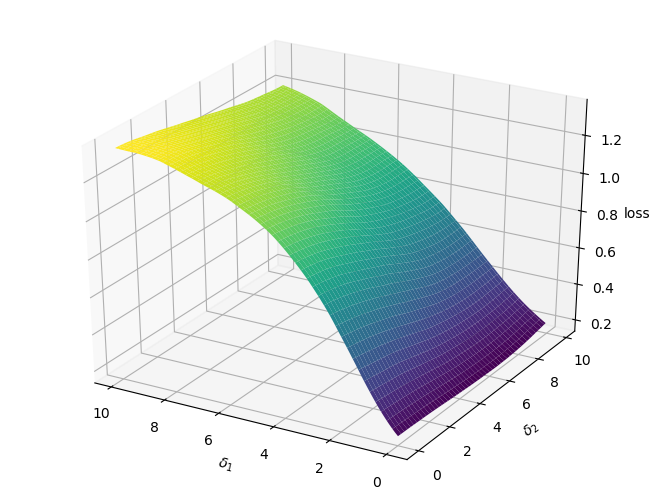}           
	\end{minipage}}
	\subfigure[]{
		\begin{minipage}{0.39\columnwidth}
			\centering                                                          
			\includegraphics[width=1\columnwidth]{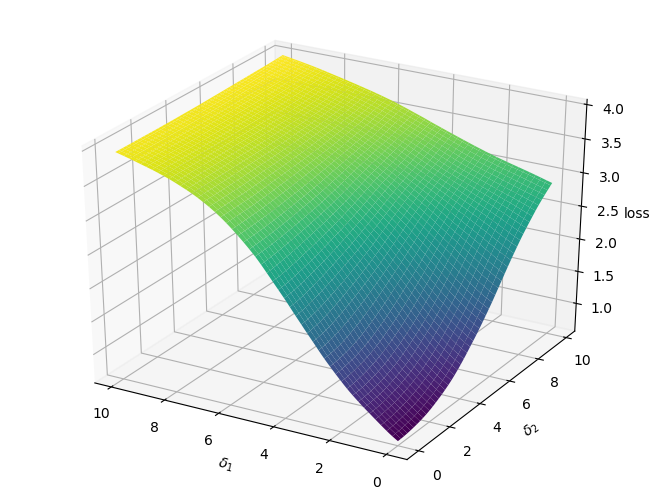}           
	\end{minipage}}
	\subfigure[]{
		\begin{minipage}{0.38\columnwidth}
			\centering                                                          
			\includegraphics[width=1\columnwidth]{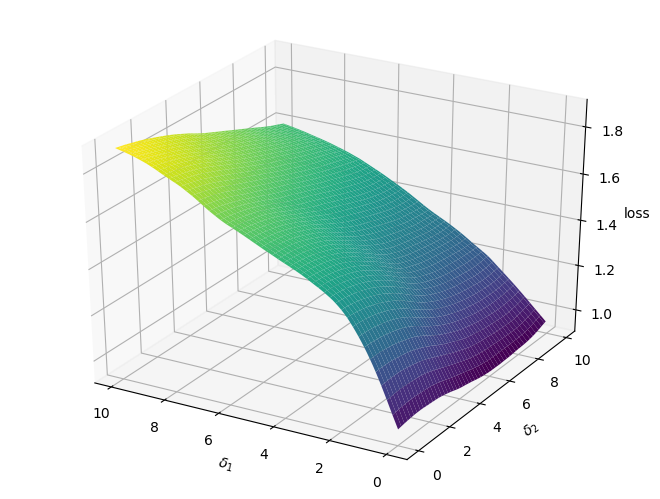}           
	\end{minipage}}
	\subfigure[]{
		\begin{minipage}{0.38\columnwidth}
			\centering                                                          
			\includegraphics[width=1\columnwidth]{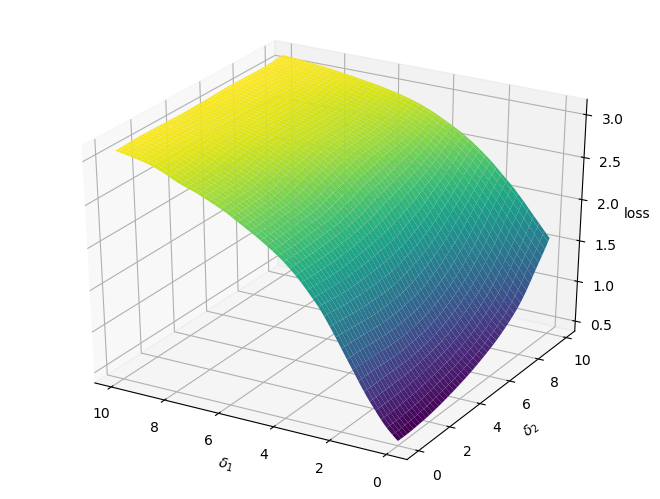}           
	\end{minipage}}
	\subfigure[]{
		\begin{minipage}{0.38\columnwidth}
			\centering                                                          
			\includegraphics[width=1\columnwidth]{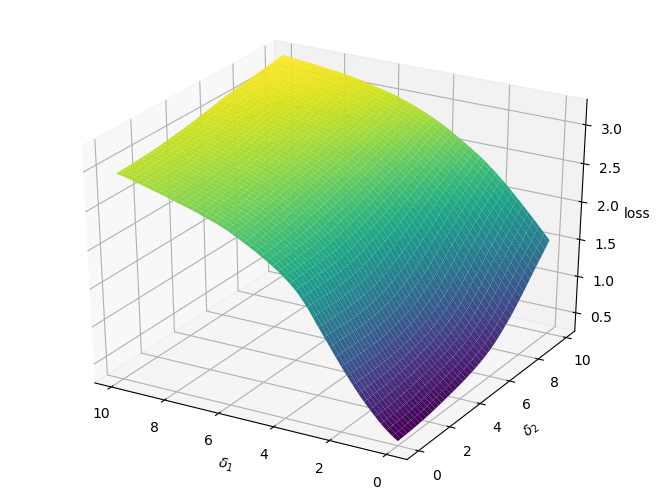}           
	\end{minipage}}
	\subfigure[]{
		\begin{minipage}{0.38\columnwidth}
			\centering                                                          
			\includegraphics[width=1\columnwidth]{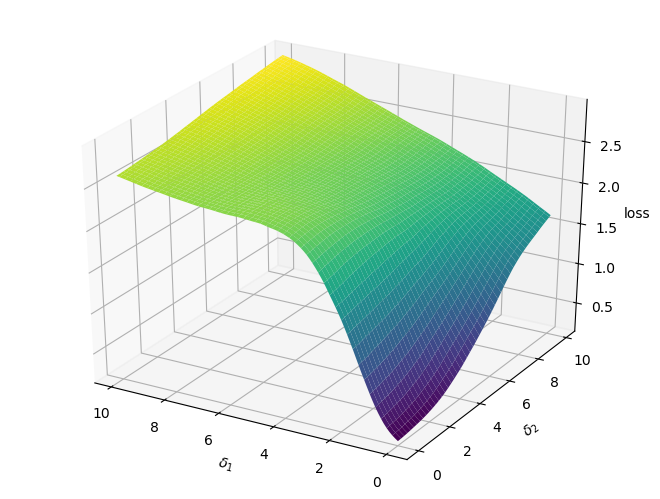}           
	\end{minipage}}
	\subfigure[]{
		\begin{minipage}{0.38\columnwidth}
			\centering                                                          
			\includegraphics[width=1\columnwidth]{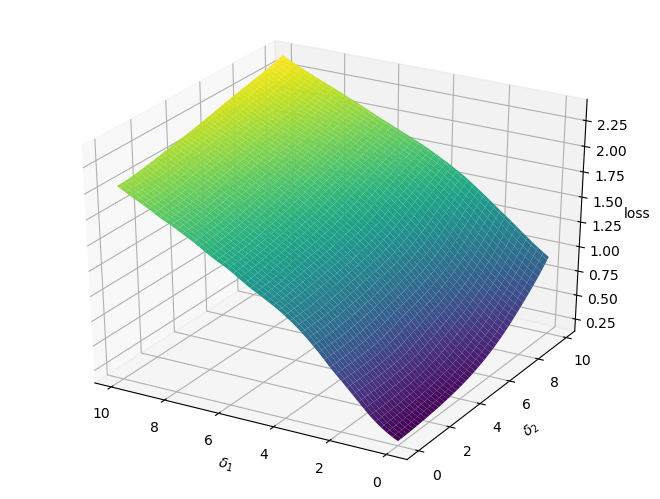}           
	\end{minipage}}
	\subfigure[]{
		\begin{minipage}{0.38\columnwidth}
			\centering                                                          
			\includegraphics[width=1\columnwidth]{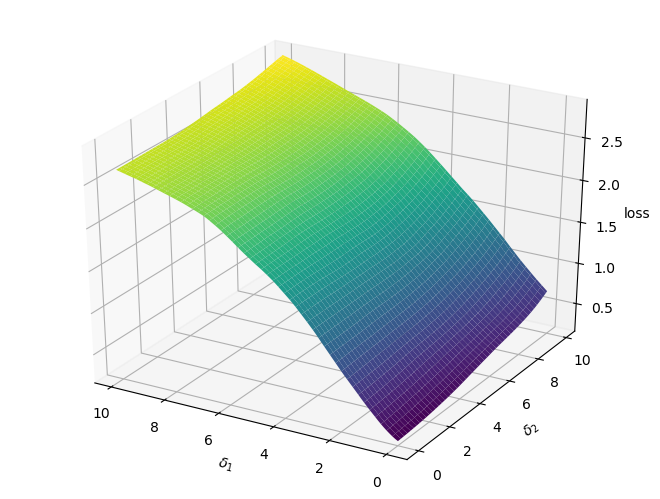}           
	\end{minipage}}
	\caption{Plot of the loss surface of an InfoAT trained model on perturbed images $\bx_i^\prime, i \in \{1,2,3,4,5,6,7,8,9,10\}$. The dataset is CIFAR-10. The model is ResNet-18.}                    
	\label{fig:app_losssurface}                                             
\end{figure*}

\begin{table*}
	\centering
	\caption{Average decision boundary (ADBD) and robust accuracy (Rob.Acc) of defense methods against black-box hard-label RayS attack.}
	\centering
	\begin{tabu}{c ccccc c cc}
		\toprule
		\multirow{3}{*}{Methods}& \multicolumn{5}{c}{CIFAR-10} & &\multicolumn{2}{c}{SVHN} \\
		\cmidrule(lr){2-6} \cmidrule(lr){8-9}
		& \multicolumn{2}{c}{ResNet18} & & \multicolumn{2}{c}{WideResNet} & & \multicolumn{2}{c}{ResNet18} \\
		\cmidrule(lr){2-3} \cmidrule(lr){5-6} \cmidrule(lr){8-9}
		& ADBD & Rob.Acc & & ADBD & Rob.Acc & & ADBD & Rob.Acc \\
		\midrule
		GanDef& 0.0328&49.60&&0.0376&55.10&&0.0367&50.00\\
		AT & 0.0361 & 53.30 & & 0.0371 & 54.20 & & 0.0355 & 49.80 \\
		TRADES & 0.0379 & 51.90 & & 0.0402 & 55.20 & & \textbf{0.0383} & 52.80 \\
		MART & 0.0374 & 52.60 & & 0.0399 & 56.50 & & 0.0374 & 53.30 \\
		InfoAT (Ours) & \textbf{0.0380} & \textbf{56.20} & & \textbf{0.0402} & \textbf{56.50} & & 0.0378 & \textbf{53.60} \\
		\bottomrule
	\end{tabu}
	\label{tab:rays_app}
\end{table*}

Then, we investigate InfoAT from four aspects: (1) Evaluating the effectiveness of surrogating mutual information with entropy, (2) employing different weighting methods, (3) replacing divergence of InfoAT, (4) exploring different regularizations in the outer minimization. 
\paragraph{The effectiveness of surrogating mutual information with entropy}
In this part, we evaluate the effectiveness of our method (i.e., surrogating mutual information with entropy) by comparing with MINE.
MINE is a method estimates the mutual information of $m \geq M$ examples. Therefore, we regularize each batch of $m$ examples with the same mutual information estimated by MINE, as shown below:
\begin{equation}
\frac{1}{n}\sum_{i=1}^{b}\sum_{j=1}^{m} I(\bx_{i:};\bz_{i:}) \mathds{1}(\htheta(\bx_{ij}^\prime) \neq \htheta(\bx_{ij})),
\end{equation}
where $b$ is the number of batch, $m$ is the number of examples in each batch,  $I(\bx_{i:};\bz_{i:})$ represents the mutual information on batch $i$. 
Here, we show the results of model robustness and accuracy based on MINE estimation mutual information on different parameter $\lambda$ in TABLE~\ref{tab:app_mutual}, where $\{\#1,\#2,\#3,\#4\}$ represent the mutual information estimated by MINE on different layer latent representations $\bz^{(l)}, l \in \{\#1,\#2,\#3,\#4\}$. The other settings are the same as the InfoAT.

The best model robustness of mutual information estimated by MINE and surrogated by entropy is shown in Fig.~\ref{fig:mu}, where $\#5$ represents the mutual information surragoted by entropy based on softmax output on the last layer $\#5$.
From Fig.~\ref{fig:mu} we can observe, surrogating mutural information with entropy (i.e., red bar) leads to a significant robustness improvement compared with any other MINE estimation mutual information.


\paragraph{Employing different weighting methods.} Instead of the weighting method which selects hard examples with high mutual information, we also evaluate the effect of other methods (i.e., unweighted (giving all examples the same weight) and misclassification aware weighting method~\cite{wang2019improving}) in Fig.~\ref{fig:weig}.
Fig.~\ref{fig:weig} shows that the robustness of model is higher by using entropy ($H(x\_{nat})$, red line) than unweighted ($None$, green line) or misclassification aware ($1-p(x\_{nat})$, orange line), which indicates the effectiveness of our claim that using mutual information (lower bounded by entropy) to select hard examples.

\paragraph{Replacing divergence of InfoAT.} As we show in Fig.~\ref{fig:diver}, when the MSE divergence (red line) is replaced by Jensen-Shannon (JS, orange line)~\cite{fuglede2004jensen}, Cross-entropy (CE, green line)~\cite{de2005tutorial}, and Kullback-Leibler divergence (KL, blue line)~\cite{van2014renyi}, the final robustness decreases by a substantial amount. It suggests that learning with MSE rather than other divergences can constrain the distribution of adversarial example with its natural example more efficiently. 

\paragraph{Exploring different regularization in the outer minimization.} Here, we show the contribution of our proposed regularization (i.e., $-I(\hat\bx_i^\prime;\hat\bz_i^\prime)$ in Equation~(\ref{equ:reg}) as well as $-H(\bp(\hat{\bx_i}^\prime))$ in Equation~(\ref{equ:ob})) to the final robustness. Specifically, we perform five regularization terms of natural examples (i.e., $+H(x\_{nat})$ with orange line, $-H(x\_{nat})$ with purple line), adversarial examples (i.e., $+H(x\_{adv})$ with green line, $-H(x\_{adv})$ with red line)  and no regularization (i.e., $None$ with blue line) in Fig.~\ref{fig:reg}. As can be observed, the robustness can be improved steadily by our proposed regularization (red line). This verifies the effectiveness of our regularization, i.e., improving the mutual information of adversarial example with its associated representation during training.

\begin{table*}[t]
	\centering
	\caption{White-box robustness (accuracy (\%) on white-box test attacks, $\epsilon=8/255$) on CIFAR-10 and CIFAR-100 with WideResNet-34-10.}
	\label{tab:wide}
	\centering
	\begin{tabu}{c|c|ccccccccc|c}
		\toprule
		Dataset & Model & Natural &FGSM& $\rm PGD^{20}$ & $\rm PGD^{100}$ & $\rm PGD^+$ & $\rm CW_\infty^{20}$ & $\rm CW_\infty^{100}$ & BPDA& $\rm InfoPGD^{20}$&AA \\
		\midrule
		\multirow{6}{*}{CIFAR-10} & GanDef& \textbf{88.78}&68.04&55.68&52.84&52.98&54.34&52.19&54.43&53.83&48.20\\
		& AT & 87.05 & 66.58 & 53.56 & 51.22 & 51.53 & 53.54 & 51.67 & 56.05&56.26&49.70 \\
		& TRADES & 84.93 & 66.95 & 56.48 & 54.72 & 54.63 & 54.61 &53.48 & 57.30&57.53&52.35 \\
		& MART & 84.10 & 67.14 & 58.29 & 56.77 & 56.67 & 54.49 & 53.02 & 57.16&57.59&51.64 \\
		& MART+ & 84.22&68.43&59.50&57.75&57.73&55.26&53.34&58.60&58.91&51.70\\
		& InfoAT (Ours) & 85.62 & \textbf{68.91} & \textbf{60.46} & \textbf{59.17} & \textbf{59.00} & \textbf{55.32} & \textbf{54.07} & \textbf{59.54}&\textbf{60.09}&\textbf{52.86} \\
		\midrule
		\multirow{6}{*}{CIFAR-100} & GanDef& \textbf{68.68}&41.04&29.17&26.36&26.64&29.85&29.55&27.84&27.66&24.05 \\
		& AT & 60.86 & 36.63 & 27.67 & 26.12 & 26.20 & 27.10 & 26.38 &32.48&32.55& 24.51 \\
		& TRADES & 56.75 & 37.91 & 31.31 & 30.41 & 30.27 & 28.22 & 27.78 & 31.91&32.20&27.09 \\
		& MART & 56.07 & 40.10 & 34.84 & 33.60 & 33.39 & 30.91 & 30.32 & 34.20&34.99&28.53 \\
		& MART+ & 53.76&40.59&35.12&34.55&34.38&29.88&29.03&34.52&35.64&28.62\\
		& InfoAT (Ours) & 60.29 & \textbf{42.85} & \textbf{35.55} & \textbf{34.89} & \textbf{34.69} & \textbf{31.37} & \textbf{30.74} &\textbf{34.37}&\textbf{37.05}&\textbf{28.99} \\
		\bottomrule 
	\end{tabu}
\end{table*}

\begin{table}
	\caption{White-box robustness (\%) on WideResNet-28-10 with 500K unlabeled data.}
	\centering
	\setlength{\tabcolsep}{5.0mm}{
		\begin{tabular}{c|cc}
			\toprule
			Model & Natural & $\rm PGD^{20}$ \\
			\midrule
			UAT++ & 86.21 &62.76 \\
			RST & 90.60 &63.60 \\
			MART & 90.20 &64.80 \\
			InfoAT (Ours) & 90.00 & \textbf{65.40}\\
			\bottomrule			
		\end{tabular}
	}
	\label{tab:semi_new}
\end{table}
\subsection{Robustness Evaluation and Analysis}
In this part, we evaluate the robustness of InfoAT on CIFAR-10, CIFAR-100 and SVHN datasets with ResNet18 against various white-box and black-box attacks. 
We consider five types of representative adversarial training methods here: 1) AT~\cite{madry2018towards}, 2) TRADES~\cite{zhang2019theoretically}, and 3) MART~\cite{wang2019improving}, 4) MART+ (replacing 1-prob in MART~\cite{wang2019improving} with the proposed mutual information weight), and 5) GanDef~\cite{liu2019gandef}, following their original papers.

\paragraph{White-Box robustness}

The robustness of all defense models is evaluated against nine types of attacks:
FGSM, $\rm PGD^{20}$, $\rm PGD^{100}$ (20-step/100-step PGD with step size $\epsilon/4$), $\rm PGD^+$ ($\rm PGD^+$ is PGD with five random restarts, and each restart has 40 steps with step size 0.01, which keeps the same as~\cite{carmon2019unlabeled}), $\rm CW_{\infty}^{20}$, $\rm CW_{\infty}^{100}$ 
($L_\infty$ version of CW optimized through $\rm PGD^{20}/ \rm PGD^{100}$), BPDA~\cite{athalye2018obfuscated}, $\rm InfoPGD^{20}$ (20-step adaptive PGD attack~\cite{tramer2020adaptive} using Equation~(\ref{equ:findadv}) to find adversarial examples),
and Auto Attack (AA)~\cite{croce2020reliable}, which is a powerful and reliable attack that integrates many parameter-free attacks to evaluate the robustness of the network. AA includes three white-box attacks (APGD-CE, APGD-DLR and FAB~\cite{croce2020minimally}) and a black-box attack (Square Attack~\cite{ andriushchenko2020square}).
All attacks are subject to the same $L_\infty$ perturbation limit, specifically, $\epsilon=8/255$.
TABLE~\ref{tab:white} shows the white-box robustness of all defense models, where “Natural” represents the accuracy on natural test images. 
From TABLE~\ref{tab:white} we can observe, InfoAT achieves the best robustness against all nine types of attacks on all CIFAR-10, CIFAR-100 and SVHN datasets.
For example, on CIFAR-10 dataset, when attacked by BPDA, InfoAT achieves 56.84\% defense accuracy, which is much higher than that (48.18\%) on GanDef. When we apply the adaptive $\rm InfoPGD^{20}$ attack, we get a accuracy of
33.92\% for InfoAT on CIFAR-100, which is still higher than the accuracy of 30.58\% for AT.
Furthermore, compared with MART, we can observe that MART+ further improves the robustness of the model (e.g., on SVHN dataset, the accuracy of MART+ is 57.15\% when attacked by $\rm CW_{\infty}^{20}$, while the accuracy of MART is 56.95\%).
This verifies the effectiveness of using mutual information to weight examples.
It should be noted that the InfoAT robustness improvement is not caused by the so-called “obfuscated gradients”~\cite{athalye2018obfuscated}. This can be verified from three phenomenons: (1) strong test attacks (e.g., $\rm PGD^{20}$) have higher success rates (lower accuracies) than weak test attacks (e.g., FGSM), and (2) InfoAT can defense AA effectively (i.e., InfoAT achieves 49.67\% accuracy when attacked by AA on CIFAR-10 dataset).
(3) Following~\cite{wu2020adversarial}, 
we plot the adversarial loss change by moving the weight $\bw$ along a random direction $d$ with magnitude $\delta$ to visualize the weight loss surface.
As presented in Fig.~\ref{fig:app_wloss}, the weight loss surface is flatter (red line) than AT (blue line), TRADES (orange line), and MART (green line). This indicates the effectiveness and stablity of our proposed InfoAT.

(4) Fig.~\ref{fig:app_losssurface} shows the loss surfaces of our proposed InfoAT trained model on perturbed images of the form $\bx_i^\prime=\bx_i+\delta_1v+\delta_2r$, obtained by varying $\delta_1$ and $\delta_2$. We craft $\rm PGD^{10}$ perturbation as the normal direction $v$, and $r$ be a random direction, under the $L_\infty$ constraint of $8/255$. From Fig.~\ref{fig:app_losssurface} we can observe, the loss surfaces of InfoAT trained model on perturbed images are smooth, which reveals the fact that the robustness improvement of our proposed InfoAT is not caused by gradient obfuscation.

\paragraph{Black-Box robustness}

We utilize SPSA~\cite{uesato2018adversarial} to perform query-based black box attacks. To estimate the gradients, we set the batch size to 128, 256, 512, and 1024, the learning rate to 0.01, and the perturbation size to 0.001. SPSA attacks are run for 100 iterations, and early-stop when it caused misclassification. 
TABLE~\ref{tab:black} shows the experiment results.
Our method obtains better robustness over other five methods in each setting. The results verify InfoAT reliably improves the robustness rather than causing gradient masking.

We further perform more black-box attack using hard-label attack RayS~\cite{chen2020rays}.
We record the Average Decision Boundary Distance (ADBD) and robust accuracy (Rob.Acc) against RayS in TABLE~\ref{tab:rays_app}. On CIFAR-10 dataset, InfoAT achieves maximum ADBD and best robust accuracy within ResNet18 and WideResNet-34-10. On SVHN, InfoAT also achieves competitive ADBD and best robust accuracy.

These analysis verify that InfoAT improves adversarial robustness effectively rather than  gradient obfuscation or masking.
\subsection{Benchmarking the State-of-the-Art Robustness}

In this part, we conduct more experiments on the large-capacity network WideResNet to arrive at a state-of-the-art robustness benchmark. In addition, we also used unlabeled data for further robustness evaluation.

\paragraph{Performance on WideResNet}  
We evaluate the robustness of our proposed InfoAT defense method more comprehensively on the WideResNet-34-10 network, and given the state-of-the-art robustness benchmarks on the CIFAR-10 and CIFAR-100 datasets. TABLE~\ref{tab:wide} shows the results of our experiment. 
As shown in the TABLE~\ref{tab:wide},
the InfoAT we proposed is superior to all baseline methods in terms of robustness. Especially under the $\rm PGD^{20}$ attack on CIFAR-10, which is the most common comparison setting. InfoAT improves $\sim 7\%$ over standard AT, $\sim 6\%$ over GanDef, $\sim4\%$ over TRADES and $\sim2\%$ even over MART. For CIFAR-100, a similar improvement trend can also be observed.

\paragraph{Boosting with additional unlabeled data} In this part, the effectiveness of proposed InfoAT is evaluated on semi-supervised version.
Following the exact settings in UAT++~\cite{uesato2019labels}, RST~\cite{carmon2019unlabeled}, and MART, we compare the robustness of InfoAT with them on WideResNet-28-10 against $\rm PGD^{20}$ respectively. The dataset is CIFAR-10 with 500K unlabeled data, which is extracted from the 80 Million Tiny Images dataset~\cite{torralba200880}. 
As confirmed in TABLE~\ref{tab:semi_new}, our proposed InfoAT also achieves the best robustness (e.g., 65.4\% accuracy on WideResNet-28-10 against $\rm PGD^{20}$ with 500K unlabeled data). This again verifies the benefit of exploiting mutual information to select ``hard" examples for improving robustness, and further 
provess the superiority of our proposed method.

\section{Conclusion}
In this paper, inspired by information bottleneck (IB) principle, we showed that the example with high mutual information is more likely to be attacked.
Based on this observation, we designed a novel and effective adversarial training using the IB principle (InfoAT), which encourages the network to find hard examples with high mutual information and guarantee them to be robust against attackers, thus improving the robustness of model.
As the experimental results show, with respect to the state-of-the-ar technology, InfoAT can significantly improve adversarial robustness, and it can also achieve the best robustness in semi-supervised task.
In the future, 
we plan to 
investigate the trade-off between robustness and natural accuracy on adversarial training, and 
exploit the different impact of examples to improve the robustness as well as accuracy of models efficiently.

 

%

\section*{Acknowledgment}
This work was supported by the National Natural Science Foundation of China (Nos. 62136004, 61876082, 61732006), the National Key R\&D Program of China (Grant Nos. 2018YFC2001600, 2018YFC2001602), and also by the CAAI-Huawei MindSpore Open Fund.

\ifCLASSOPTIONcaptionsoff
  \newpage
\fi




\bibliographystyle{IEEEtran}
\bibliography{reference}
\end{document}